# Distance-Geometric Graph Convolutional Network (DG-GCN) for Three-Dimensional (3D) Graphs

Daniel T. Chang (张遵)

*IBM (Retired)* dtchang43@gmail.com

**Abstract:**

The distance-geometric graph representation adopts a unified scheme (distance) for representing the geometry of three-dimensional (3D) graphs. It is invariant to rotation and translation of the graph and it reflects pair-wise node interactions and their generally local nature. To facilitate the incorporation of geometry in deep learning on 3D graphs, we propose a message-passing graph convolutional network based on the distance-geometric graph representation: DG-GCN (distance-geometric graph convolution network). It utilizes continuous-filter convolutional layers, with filter-generating networks, that enable learning of filter weights from distances, thereby incorporating the geometry of 3D graphs in graph convolutions. Our results for the ESOL and FreeSolv datasets show major improvement over those of standard graph convolutions. They also show significant improvement over those of geometric graph convolutions employing edge weight / edge distance power laws. Our work demonstrates the utility and value of DG-GCN for end-to-end deep learning on 3D graphs, particularly molecular graphs.

## 1 Introduction

The *geometry* of three-dimensional (3D) graphs, consisting of nodes and edges, plays a crucial role in many important applications. An excellent example is molecular graphs, whose geometry influences important properties of a molecule including its reactivity and biological activity. As in [1], we focus on 3D graphs whose geometry can be fully specified in terms of *edge distances (d), angles (θ) and dihedrals (φ)*. A key advantage of such specification is its *invariance to rotation and translation* of the graph.

*Distance geometry* [2-3] is the characterization and study of the geometry of 3D graphs based only on given values of the *distances* between pairs of nodes. From the perspective of distance geometry, the geometry of 3D graphs can be equivalently specified in terms of *edge distances (d), angle distances ($d^θ$) and dihedral distances ($d^φ$)*. In addition to *invariance to rotation and translation* of the graph, such specification adopts a *unified scheme (distance)* and reflects *pair-wise node interactions* and their generally local nature.

To facilitate the incorporation of geometry in deep learning on 3D graphs, three types of *geometric graph representations* are defined in [1]: positional, angle-geometric and distance-geometric. The *positional graph representation* is based on node positions, i.e., Cartesian coordinates of nodes. The *angle-geometric graph representation* centers on edge distances (d), angles (θ) and dihedrals (φ); it is invariant to rotation and translation of the graph. The *distance-geometric*

*graph representation* is based on *distances*: edge distances (d), angle distances ($d^\theta$) and dihedral distances ($d^\varphi$); it is invariant to rotation and translation of the graph and it reflects pair-wise node interactions and their generally local nature.

*Graph convolutional networks (GCNs)* [7] have been applied to deep learning on graphs. However, standard GCNs do not take spatial arrangements of the nodes and edges into account. Therefore, they can accommodate only graph constitution, but not graph geometry. To incorporate geometry in graph convolutions, g*eometric graph convolutions* [1] use the distance-geometric graph representation and employ edge weight / edge distance power laws. The combination enables the incorporation of geometry in graph convolutions utilizing standard GCNs by (1) expanding the kinds of edges involved to include not just *edges (e)* with neighbor nodes, but also *angle edges ($e^\theta$)* with second-order-neighbor nodes and *dihedral edges ($e^\varphi$)* with third-order-neighbor nodes, and (2) assigning different *weight*s to different edges based on their kind and their distance.

We take the step further and propose *DG-GCN (distance-geometric graph convolutional network)*, a message-passing graph convolutional network based on the distance-geometric graph representation. Similar to geometric graph convolutions, DG-GCN considers all edges that are important to graph geometry in graph convolutions. These include *(connected) edges*, *angle edges* and *dihedral edges*.

However, instead of using hand-crafted edge weight / edge distance power laws, DG-GCN utilizes *continuous-filter convolutional layers* [4-5], with *filter-generating networks*, which enable learning of filter weights from distances. This enables end-to-end deep learning on 3D graphs.

DG-GCN is implemented using *PyTorch Geometric (PyG)* [6]. In particular, the implementation adopts *CFConv*, extracted from the schnet module in PyG, as the continuous-filter convolutional layer.

Review of the related work that are referenced, but not discussed, in the main body of the paper is provided in the Appendix. So are additional results for the QM9 dataset.

## 2 Geometry of 3D Graphs

The *geometry* of 3D graphs is the three-dimensional arrangement of the *nodes* and *edges* in a graph. It is often specified in terms of the *Cartesian coordinates of nodes*. However, such specification depends on the (arbitrary) choice of origin and is too general for specifying geometry.

We focus on 3D graphs whose geometry can be fully specified in terms of *edge distances (d), angles (θ)* and *dihedrals (φ)*. Molecular graphs are excellent examples of such graphs. The edge distance is the distance between two nodes connected together. The angle is the angle formed between three nodes across two edges. For three edges connected in a chain, the dihedral is the angle between the plane formed by the first two edges and the plane formed by the last two edges. These are illustrated in the following diagram:

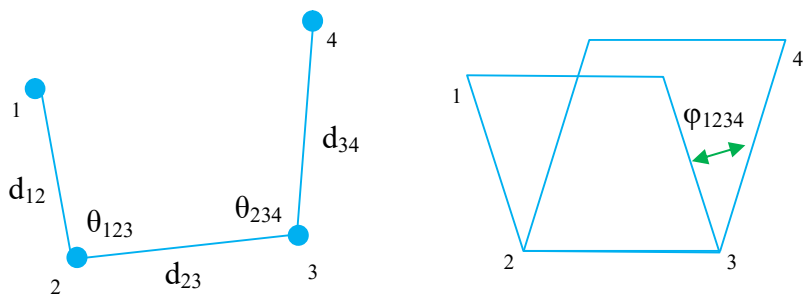

A key advantage of using edge distances, angles and dihedrals to specify geometry is its *invariance to rotation and translation* of the graph.

## 2.1 Distance Geometry of 3D Graphs

*Distance geometry* [2-3] refers to a foundation of geometry based on the concept of *distances* instead of those of points and lines or point coordinates. For 3D graphs, distance geometry is the characterization and study of their geometry based only on given values of the *distances* between pairs of nodes. From the perspective of distance geometry, the geometry of 3D graphs can be equivalently specified in terms of *edge distances (d), angle distances ($d^\theta$)* and *dihedral distances ($d^\varphi$)*. The angle distance is the distance of the *angle edge ($e^\theta$)* and the dihedral distance is the distance of the *dihedral edge ($e^\varphi$)*. The angle edge is the unconnected, end edge between the end nodes of an angle and the dihedral edge is the unconnected, end edge between the end nodes of a dihedral. (We therefore refer to both angle edges and dihedral edges as *end edges*.) These are illustrated in the following diagram (with dashed lines representing angle edges and dotted lines representing dihedral edges):

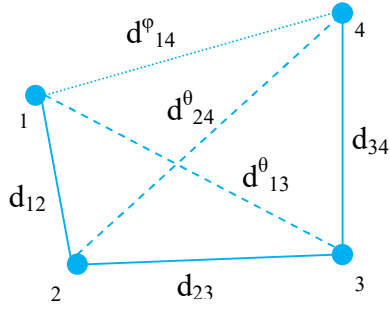

As the case of using edge distances, angles and dihedrals to specify geometry, a key advantage of specifying geometry in terms of edge distances, angle distances and dihedral distances is its *invariance to rotation and translation*. In addition, it adopts a *unified scheme (distance)* and reflects *pair-wise node interactions* and their generally local nature, which are additional advantages.

## 3 Geometric Graph Representations

To facilitate the incorporation of geometry in deep learning on 3D graphs, three types of *geometric graph representations* are defined in [1]: positional, angle-geometric and distance-geometric. The *positional graph representation* is based on node positions, i.e., Cartesian coordinates of nodes. The *angle-geometric graph representation* centers on edge distances (d), angles (θ) and dihedrals (φ); it is invariant to rotation and translation of the graph. The *distance-geometric graph representation* is based on *distances*: edge distances (d), angle distances ($d^\theta$) and dihedral distances ($d^\varphi$); it is invariant to rotation and translation of the graph and it reflects pair-wise node interactions and their generally local nature.

There are recent work on deep learning on 3D graphs that use the position graph representation [9-10] and the angle-geometric graph representation [11-12]. As in [1], we focus on using the distance-geometric graph representation due to its advantages, uniformity and simplicity. For convenience of discussion, we repeat the definition of the distance-geometric graph representation in the following.

### 3.1 Distance-Geometric Graph Representation

From the perspective of distance geometry, a 3D graph can be represented as G = (**X**, (**I**, **E**), (**D**, $\mathbf{D}^\theta$, $\mathbf{D}^\varphi$)) where **X** ∈ $R^{N \times d}$ is the *node feature matrix* and (**I**, **E**) is the *sparse adjacency tuple*. **I** ∈ $N^{2 \times U}$ encodes *edge indices* in coordinate (COO) format and **E** ϵ $R^{U \times s}$ is the *edge feature matrix*. **D** ϵ $R^{U \times 1}$ is the *edge distance matrix*, $D^\theta$ ϵ $R^{U^\theta \times 1}$ is the *angle distance matrix*, and $D^\varphi$ ϵ $R^{U^\varphi \times 1}$ is the *dihedral distance matrix*. U is the number of edges, $U^\theta$ is the number of angles, and $U^\varphi$ is the number of dihedrals.

This representation is based on distances; therefore we refer to it as the *distance-geometric graph representation*. As discussed in Section 2.1, it is *invariant to rotation and translation* of the graph. In addition, it reflects *pair-wise node interactions* and their generally local nature. These are very useful for *graph convolutions*, which are locally oriented. They are particular useful for *molecular graphs*, since electrostatic, intermolecular, and other conformation-driven properties of molecules depend on the pair-wise interatomic (internodal) distances.

## 4 Distance-Geometric Graph Convolutional Network (DG-GCN)

*Graph convolutional networks (GCNs)* [7] have been applied to deep learning on graphs. However, standard GCNs do not take spatial arrangements of the nodes and edges into account. Therefore, they can accommodate only graph constitution, but not graph geometry.

To incorporate geometry in graph convolutions, g*eometric graph convolutions* [1] use the distance-geometric graph representation and employ edge weight / edge distance power laws. The combination enables the incorporation of geometry in graph convolutions utilizing standard GCNs by (1) expanding the kinds of edges involved to include not just *edges (e)* with neighbor nodes, but also *angle edges ($e^\theta$)* with second-order-neighbor nodes and *dihedral edges ($e^\varphi$)* with third-order-neighbor nodes, and (2) assigning different *weight*s to different edges based on their kind and their distance.

We take the step further and propose *DG-GCN (distance-geometric graph convolutional network)*, a message-passing graph convolutional network based on the distance-geometric graph representation.

### 4.1 Graph Convolutional Networks (GCNs)

As in [1], we start with *graph neural networks* (*GNNs*) that employ the following *message passing* scheme for node i at layer k:

$$x_i^{(k)} = \gamma^{(k)}(x_i^{(k-1)}, \Pi_j \lambda^{(k)}(x_i^{(k-1)}, x_j^{(k-1)}, e_{ij}))$$

where j ϵ N(i) denotes a neighbor node of node i. $x_i$ is the node feature vector and $e_{ij}$ is the edge feature vector. γ and λ denote differentiable update and message functions, respectively, and Π denotes a differentiable aggregation function.

The *standard GCN* [7] implements message passing using the adjacency matrix **A**:

$$\mathbf{X}^{(k)} = \check{\mathbf{D}}^{-1/2}\check{\mathbf{A}}\check{\mathbf{D}}^{-1/2}\mathbf{X}^{(k-1)}\mathbf{W}^{(k-1)},$$

where $\check{\mathbf{A}} = \mathbf{A} + \mathbf{I}$ denotes the adjacency matrix with inserted self-loops and $\check{D}_{ii}=\sum_{j=0}\check{A}_{ij}$ its diagonal degree matrix. $A_{ij}$ is one when there is an edge from node i to node j, and zero when there is no edge. $\mathbf{W}^{(k-1)}$ is the layer-specific weight matrix.

In the context of message-passing GNNs, the standard GCN takes the following shape: [8]

$$\mathbf{x}_i^{(k)} = \frac{1}{c_i}(\mathbf{x}_i^{(k-1)} + \Sigma_j \mathbf{x}_j^{(k-1)})\mathbf{W}^{(k-1)}$$

where $c_i$ is a node-specific normalization constant.

In case the graph has edge weights, $w_{ij}$, the above equation can be expanded as:

$$\mathbf{x}_i^{(k)} = \frac{1}{c_i}(\mathbf{x}_i^{(k-1)} + \Sigma_j\, w_{ij}\mathbf{x}_j^{(k-1)})\mathbf{W}^{(k-1)}$$

This is used by *geometric graph convolutions* in [1] with $w_{ij}$ determined from edge weight / edge distance power laws and N(i) expanded as N(i) = $\mathcal{N}(i) + \mathcal{N}^\theta(i) + \mathcal{N}^\varphi(i)$, with $\mathcal{N}(i)$ being the *first-order (1st)* neighbors, $\mathcal{N}^\theta(i)$ the *second-order (2nd)* neighbors and $\mathcal{N}^\varphi(i)$ the *third-order (3rd)* neighbors.

## 4.2 DG-GCN

Similar to geometric graph convolutions, *DG-GCN* considers all edges that are important to graph geometry in graph convolutions. These include *(connected) edges (e)* with 1st neighbors, *angle edges ($e^\theta$)* with 2nd neighbors, and *dihedral edges ($e^\varphi$)* with 3rd neighbors. The following diagram [12] shows the neighborhood of a node of a molecular graph where a) includes 1st neighbors (black), b) includes 1st neighbors and 2nd neighbors (blue), and c) includes 1st neighbors, 2nd neighbors and 3rd neighbors (red). It can be seen that DG-GCN, i.e., case c), fully captures the local geometry (and substructures) of a node in graph convolutions.

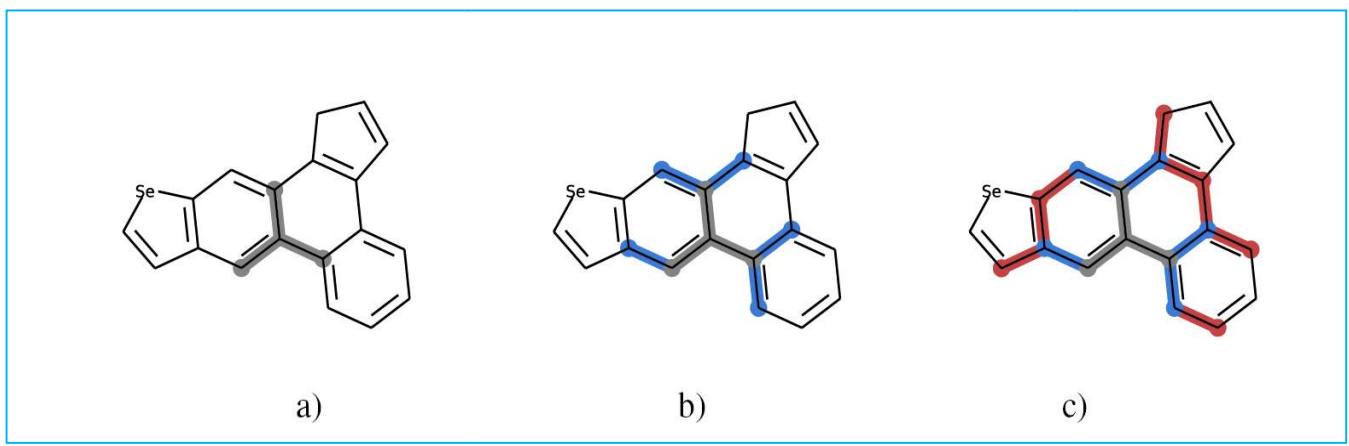

However, instead of using hand-crafted edge weight / edge distance power laws, DG-GCN utilizes *continuous-filter convolutional layers* [4-5], with *filter-generating networks*, which enable learning of filter weights from distances. This enables end-to-end deep learning on 3D graphs.

The architecture of DG-GCN is shown below, where $\{d_{ij}\}$ denotes the set of all distances, including edge distances (d), angle distances ($d^\theta$) and dihedral distances ($d^\varphi$). CFConv is discussed in the next section. We use mean pooling to be consistent with previous studies; sum pooling, however, produces better results as shown in Section 5.

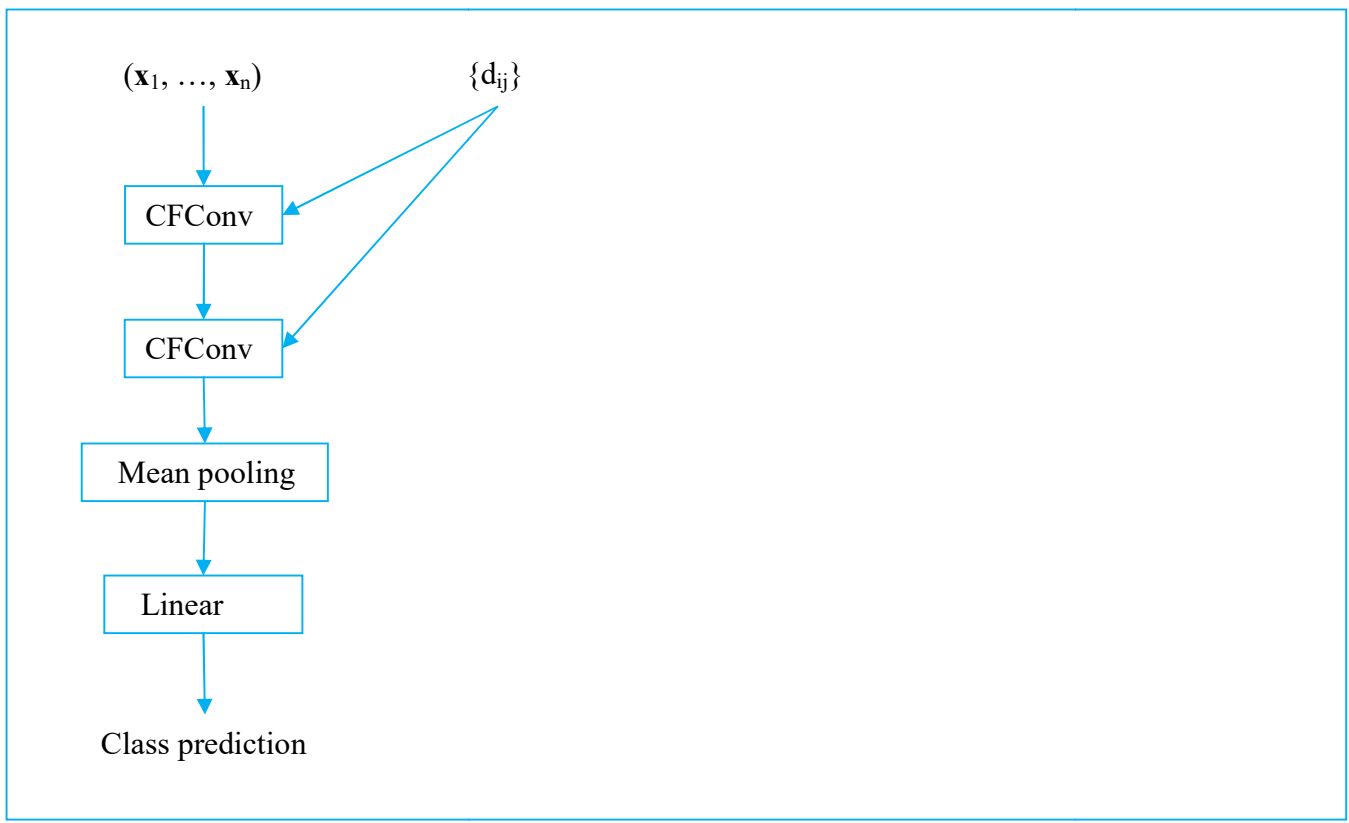

## 4.3 Continuous-Filter Convolutional Layer (CFConv)

The *continuous-filter convolutional layer (CFConv)* [2-3] is a generalization of the discrete convolutional layers commonly used for image or audio data. This generalization is necessary since nodes in a 3D graph are not located on a regular grid like pixels, but can be located at arbitrary positions.

The filters are modeled in a continuous fashion with a *filter-generating network (FGNet)*, to be discussed next, that maps from distances, $d_{ij}$, to corresponding filter weights, $W^{(k)}(d_{ij})$, at layer k. The output for the convolutional layer is then given by

$$\mathbf{x}_i^{(k)} = \frac{1}{c_i}(\mathbf{x}_i^{(k-1)} + \Sigma_j w_{ij} \mathbf{x}_j^{(k-1)} W^{(k-1)}(d_{ij}))$$

where $j \in N(i)$ denotes the set of neighbors of node i. $N(i) = \mathcal{N}(i) + \mathcal{N}^\theta(i) + \mathcal{N}^\varphi(i)$, with $\mathcal{N}(i)$ being the 1st neighbors, $\mathcal{N}^\theta(i)$ the 2nd neighbors and $\mathcal{N}^\varphi(i)$ the 3rd neighbors. $w_{ij}$ is determined from $d_{ij}$ using $w_{ij} = 0.5 * \cos((d_{ij} / d_{cutoff}) * \pi) + 1)$ where $d_{cutoff}$ is the distance cutoff (see Section 4.4). $w_{ij}$ varies from 1 ($d_{ij} = 0$) to 0 ($d_{ij} = d_{cutoff}$).

The architecture of CFConv is shown below:

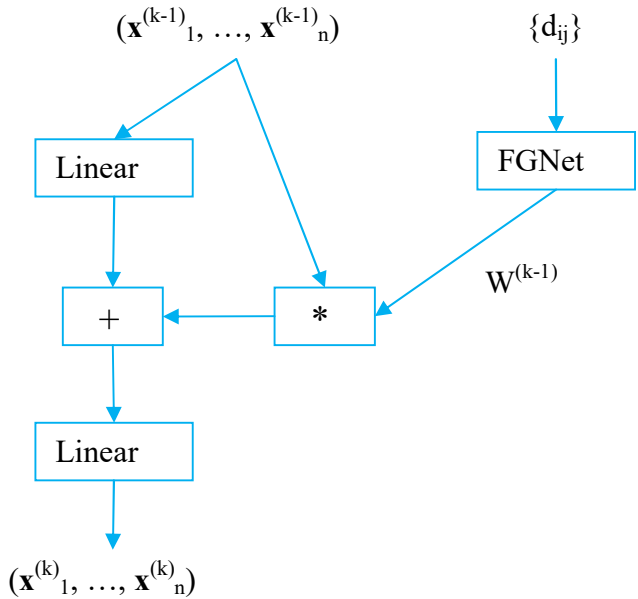

## 4.4 Filter-Generating Network (FGNet)

The *filter-generating network (FGNet)* [2-3] learns *filter weights* from *distances*. It is a fully-connected network and it is rotational invariant since it uses distances as input. The distance, $d_{ij}$, is first expanded in a basis of Gaussians

$$e_k(d_{ij}) = \exp(-\gamma(d_{ij} - \mu_k)^2)$$

with *centers* $\mu_k$ located on a uniform grid between zero and the *distance cutoff*. Due to this additional non-linearity, the initial filters are less correlated leading to a faster training procedure. The *number of Gaussians* and the *hyperparameter γ* determine the resolution of the filter. For the feasibility study, we use the default values provided in SchNet [2-3] for the number of Gaussians, distance cutoff, $\mu_k$ and γ.

The architecture of FGNet is shown below. *ShiftedSoftPlus* is defined as $ssp(x) = \ln(0.5e^x + 0.5)$. The shifting ensures that $ssp(0) = 0$ and improves the convergence of the network while having infinite order of continuity.

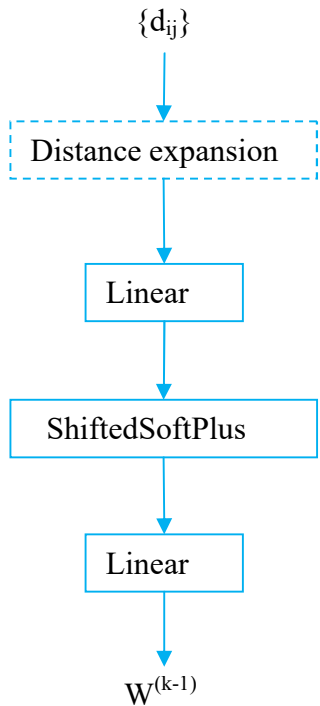

# 5 Experiments

A number of experiments have been carried out using the *ESOL* and *FreeSolv* datasets, which are used in [9-10] for training and evaluating 3D-extended GCNs. Specifically, dataset files provided by Geo-GCN [10] are used, same as [1], which contain molecular graph data including three-dimensional node coordinates. These are small datasets with *901 / 113 / 113* and *510 / 65 / 64* training / test / validation samples, respectively. Our focus, however, is on qualitatively comparing results of DG-GCN with those of standard graph convolutions and geometric graph convolutions, reported in [1], all based on the same sample sizes. That is, our interest is on relative accuracy not absolute accuracy.

The results are listed in the following table. *Standard GC* (graph convolutions) [1] utilizes the default GCNConv with all edges having a weight of one and serves as the *baseline* for comparison. *Geometric GC* [1] utilizes the GCNConv with edge weights calculated from edge distances using power laws. In particular, *Geometric GC (Ref)* denotes the reference geometric GC which uses fixed $R_0 = 1.39$ and $N = 4.55$ for the power law parameters.

As in [1], we consider three cases of geometric graph convolutions for Geometric GC and DG-GC: edges (*1$^{st}$ Nbrs*) which include 1$^{st}$ neighbor nodes, edges + angle edges (*2$^{nd}$ Nbrs*) which include 1$^{st}$ and 2$^{nd}$ neighbor nodes, and edges + angle edges + dihedral edges (*3$^{rd}$ Nbrs*) which include 1$^{st}$, 2$^{nd}$ and 3$^{rd}$ neighbor nodes. We include the 1$^{st}$ Nbrs and 2$^{nd}$ Nbrs cases to verify and show the consistency of DG-GCN results. The (full) DG-GCN results are represented by the *3$^{rd}$ Nbrs* cases.

| Dataset | Model | RMSE |
|---|---|---|
| **ESOL** | Standard GC | 0.4573 |
| | Geometric GC (Ref) – 3$^{rd}$ Nbrs | 0.4273 |
| | DG-GCN – 1$^{st}$ Nbrs | 0.3623 |
| | DG-GCN – 2$^{nd}$ Nbrs | 0.3216 |
| | DG-GCN – 3$^{rd}$ Nbrs | 0.3379 |
| | (Sum pooling) | 0.3231 |
| **FreeSolv** | Standard GC | 0.4183 |
| | Geometric GC (Ref) – 3$^{rd}$ Nbrs | 0.3710 |
| | DG-GCN – 1$^{st}$ Nbrs | 0.3608 |
| | DG-GCN – 2$^{nd}$ Nbrs | 0.3468 |
|  | DG-GCN – 3<sup>rd</sup> Nbrs | 0.3405 |
|  | (Sum pooling) | 0.2890 |

It can be seen from the table, for both the ESOL and FreeSolv datasets, the results of *DG-GCN* show major improvement over those of Standard GC. They also show significant improvement over Geometric GC. This demonstrates the utility and value of DG-GCN for end-to-end deep learning on 3D graphs, particularly molecular graphs.

## 6 Summary and Conclusion

To facilitate the incorporation of geometry in deep learning on 3D graphs, we propose a message-passing graph convolutional network based on the distance-geometric graph representation: DG-GCN (distance-geometric graph convolution network). It utilizes continuous-filter convolutional layers, with filter-generating networks, that enable learning of filter weights from distances, thereby incorporating the geometry of 3D graphs in graph convolutions.

Our results for the ESOL and FreeSolv datasets show major improvement over those of standard graph convolutions. They also show significant improvement over those of geometric graph convolutions employing edge weight / edge distance power laws.

Our work demonstrates the utility and value of DG-GCN for end-to-end deep learning on 3D graphs, particularly molecular graphs.

**Acknowledgement:** Thanks to my wife Hedy (期芳) for her support.

## Appendix 1 Related Work

**L-GCN** [8] provides a learning mechanism that transforms edge attributes into latent representations, which can then serve as input to a GCN-like architecture for further propagation in the form of an adjacency tensor. Edges are represented by a vector $w_{ij}$ *containing multiple weights* across different edge attributes. L-GCN turns the edge weights into *trainable parameters* in an end-to-end fashion, using the output of the learning mechanism which operates on either a vector describing multiple edge attributes or a sequence of such vectors (multi-edges). In contrast, DG-GCN provides a learning mechanism, through FGNet as part of CFConv, that transforms (edge) distances into filter weights in an end-to-end fashion.

**3DGCN** [9] builds a three-dimensional (3D) graph convolutional network by augmenting the standard GCN layer with the *relative (node) position matrix*, which contains the full spatial topology of a 3D graph. Specifically, it incorporates node-level *vector features*, as well as conventional *scalar features*, and brings them together by using the *interconverting operations* based on the relative position matrix for 3D graph convolutions. The relative position matrix ensures translational invariance. 3DGCN is based on the positional graph representation, though using relative positions; in contrast, DG-GCN is based on the distance-geometric graph representation.

**GeoGCN** [10] uses geometric features (spatial coordinates) in GCNs and is a proper generalization of GCNs and *CNNs (convolutional neural networks)*. The *relative positions* in the neighborhood of a node are transformed using a linear operation combined with non-linear ReLU function. This scalar is used to *weigh* the feature vectors in a neighborhood. GeoGCN is based on the positional graph representation, though using relative positions; in contrast, DG-GCN is based on the distance-geometric graph representation.

**DimeNet** [11] embeds the messages passed between nodes such that each message is associated with a *direction in coordinate space* and are rotationally equivariant since the associated directions rotate with the graph. The message passing scheme transforms messages based on the *angle between nodes* in order to encode direction. This is done by using spherical Bessel functions and spherical harmonics to construct theoretically well-founded, orthogonal representations. DimeNet is based on the angle-geometric graph representation, though considering only angles but not dihedrals; in contrast, DG-GCN is based on the distance-geometric graph representation.

**Path GCN** [12] generalizes GNNs to pass messages and aggregate across *higher order paths*. This allows for information to propagate over various levels and substructures of the graph. Specifically, Path GCN learns representations over larger node neighborhoods within each propagation layer by simply augmenting the message function to aggregate over *additional neighbors*. By summing over additional neighbors it enables the use of path features such as *angles* for paths of length two and *dihedrals* for paths of length three. Path GCN is inspired by the angle-geometric graph representation and includes neighbors up to *path length l* in graph convolutions; in contrast, DG-GCN is based on the distance-geometric graph representation and includes all, but only, neighbors germane to the graph geometry: $1^{st}$ neighbors, $2^{nd}$ neighbors and $3^{rd}$ neighbors.

# Appendix 2 Results for QM9

The results are obtained using sum pooling (see Section 4.2 and Section 5). It can be seen that among the twelve properties, the following five exhibit geometric effect: mu, LUMO, gap, $R^2$ and Cv.

| QM9 Molecular Property | Model | RMSE (Relative) |
|---|---|---|
| mu (dipole moment) | Standard GC | 0.8965 |
|  | DG-GCN | 0.6425 |
| alpha (isotropic polarizability) | Standard GC | 0.0257 |
|  | DG-GCN | 0.0219 |
| HOMO (highest occupied molecular orbital energy) | Standard GC | 0.0596 |
|  | DG-GCN | 0.0646 |
| LUMO (lowest unoccupied molecular orbital energy) | Standard GC | 1.7978 |
|  | DG-GCN | 1.4570 |
| gap (gap between HOMO and LUMO) | Standard GC | 0.1076 |
|  | DG-GCN | 0.0863 |
| $R^2$ (electronic spatial extent) | Standard GC | 0.1601 |
|  | DG-GCN | 0.0925 |
| ZPVE (zero point vibrational energy) | Standard GC | 0.0101 |
| | DG-GCN | 0.0106 |
| $U_0$ (internal energy at 0K) | Standard GC | 0.0095 |
| | DG-GCN | 0.0262 |
| U (internal energy at 298.15K) | Standard GC | 0.0095 |
| | DG-GCN | 0.0262 |
| H (enthalpy at 298.15K) | Standard GC | 0.0095 |
| | DG-GCN | 0.0262 |
| G (free energy at 298.15K) | Standard GC | 0.0095 |
| | DG-GCN | 0.0262 |
| Cv (heat capacity at 298.15K) | Standard GC | 0.0416 |
| | DG-GCN | 0.0250 |